\documentclass[a4paper, twocolumn]{revtex4-1}  
\usepackage{amssymb,latexsym,amsmath,amscd}
\usepackage{xcolor}

\usepackage{soul}

\usepackage{enumitem}

\usepackage{float}
\usepackage{hyperref}

\newcommand{\x}{{\bf x}}

\usepackage[english]{babel}
\usepackage{graphicx}
\usepackage{verbatim}

\graphicspath{{figures/}}

\definecolor{royalblue}{RGB}{4,51,255}
\definecolor{eggplant}{RGB}{180,33,147}

\begin{document}

\title{Generalization and Overfitting in Matrix Product State Machine Learning Architectures}
\date{\today}

\author{Artem Strashko}
\email{astrashko@flatironinstitute.org}
\affiliation{Center for Computational Quantum Physics, Flatiron Institute, 162 5th Avenue, New York, NY 10010, USA}

\author{E.\ Miles Stoudenmire}
\email{mstoudenmire@flatironinstitute.org}
\affiliation{Center for Computational Quantum Physics, Flatiron Institute, 162 5th Avenue, New York, NY 10010, USA}

\begin{abstract}
While overfitting and, more generally, double descent are ubiquitous in machine learning, increasing the number of parameters of the most widely used tensor network, the matrix product state (MPS), has generally lead to monotonic improvement of test performance in previous studies. To better understand the generalization properties of architectures parameterized by MPS, we construct artificial data which can be exactly modeled by an MPS and train the models with different number of parameters. We observe model overfitting for one-dimensional data, but also find that for more complex data overfitting is less significant, while with MNIST image data we do not find any signatures of overfitting. We speculate that generalization properties of MPS depend on the properties of data: with one-dimensional data (for which the MPS ansatz is the most suitable) MPS is prone to overfitting, while with more complex data which cannot be fit by MPS exactly, overfitting may be much less significant. 
\end{abstract}

\maketitle

\section{Introduction}

\begin{figure}
\includegraphics[width=0.5\linewidth]{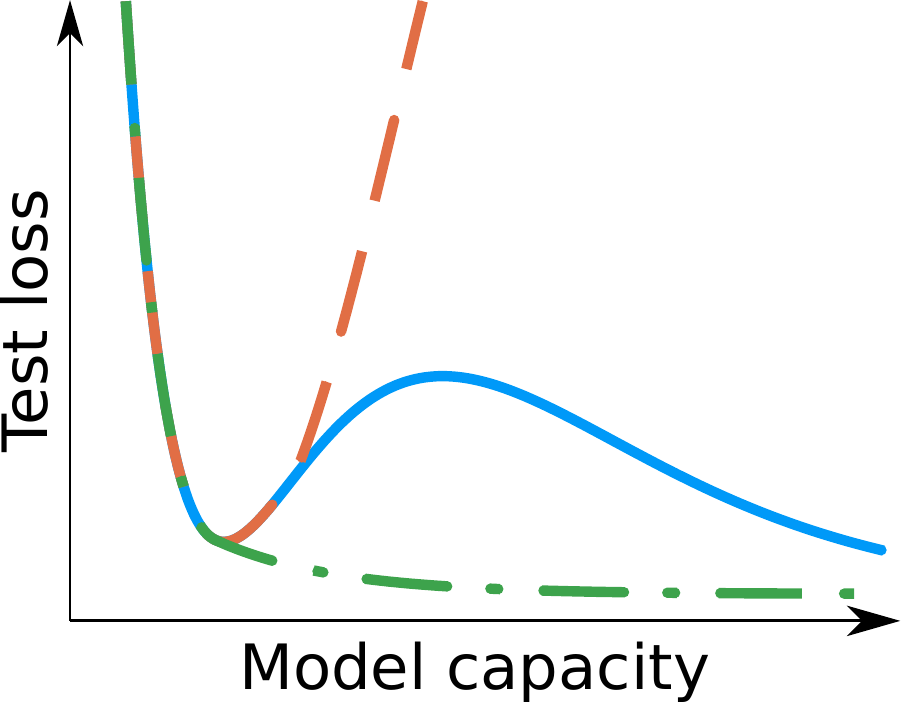}
\caption{Possible scenarios of test loss dependence on a model capacity. 
        Orange dashed line: ``classical" statistics where improving test performance 
        is followed by overfitting (U-shaped behavior). 
        Green dash--dotted line: training with regularization, which generally leads to improved 
        test performance as model capacity grows. 
        Blue solid line: double descent test loss profile (no explicit regularization) generally found in 
        most of machine learning models --- see Ref.~\cite{double_descent_Sutskever} and references therein. 
        }
\label{losses_cartoon}
\end{figure}

One of the central tasks of machine learning is to generalize model performance to unseen data. 
Classical statistical theory suggests that increasing the complexity of a model reduces its bias and 
increases its variance \cite{classical_stats} leading to the well known non-monotonic U-shaped behavior of test loss --- 
see the orange dashed line in Fig.~\ref{losses_cartoon}. 
This means that one observes either underfitting (high bias, 
low variance) at smaller model complexity or overfitting (high variance, low bias) at higher 
complexity. But modern practice suggests the reality is more complex. When training 
a model long enough without any regularization, one finds a ubiquitous double descent behavior 
of test loss \cite{double_descent_Sutskever}: a U-shaped loss profile is followed by a monotonic 
decrease of the loss as model complexity grows --- see dash--dotted green line in Fig.~\ref{losses_cartoon}. 
This behavior was observed in many different 
architectures and different domains which suggests that more complex models generalize better.

A common and powerful model used in physics to approximate probability distributions of complex 
systems in exponentially high dimensional spaces is a matrix product state (MPS) 
\cite{schollwock2011density}, also known as a tensor train \cite{oseledets2011tensor}. 
The expressivity of an MPS is controlled by its \emph{bond dimension}, or dimension of the internal
indices summed over within the network of tensors.
At higher bond dimension, an MPS is able to capture longer-ranged and more complex correlations, 
thus resolving more details of an underlying probability distribution. 

Following its success in physics applications, the MPS format has been investigated as a model
architecture for machine learning tasks such as supervised learning \cite{Exp_machines,stoudenmire2016supervised,Glasser:2018,Dborin:2021,Wright_Deterministic,Dilip:2022} and generative modeling \cite{PhysRevX.8.031012,PhysRevB.99.155131,Bradley_2020,Liu:2021,Hur:2022,Nunez:2022,MPS_without_optimization,MPS_generative_finance}.
In machine learning applications MPS and other tensor network architectures typically show 
improved performance as their capacity grows, determined by their bond dimension \cite{stoudenmire2016supervised}.
But overfitting has also been observed \cite{block_MPS,PEPS_classification}. 
This raises a question: how does increasing MPS bond dimension affect its 
generalization properties, that is, its performance on unseen test data when using finite training data? 

To address this question in a controlled way, we focus on a regression task using a complex artificial 
data set which nevertheless can be fit exactly by an MPS. By employing two different methods to obtain trained models
--- either inversion and compression or gradient--based optimization --- we observe that a finite MPS 
bond dimension acts to regularize the model. When the number of training samples is not sufficient to fully specify all tensor 
elements and when the data is not too complex (can be well represented by a low order polynomial),
there exists an optimal bond dimension $\chi^*$ which 
results in the best test performance and which is smaller than the bond dimension required to 
fit the data exactly. We also see that the optimal bond dimension $\chi^*$ generally increases as 
as the complexity of underlying data distribution grows or as the number of training samples 
goes up (increasing the amount of signal in the training set). 

The rest of the paper is organized as follows. Section~\ref{details_of_cals} describes the techniques used: 
the MPS regression approach, data construction procedure, and MPS optimization methods. 
In Section~\ref{results_with_artificial_data} we present our studies based on artificial data for which we observe 
an overfitting phenomenon.
In Section~\ref{results_with_MNIST} we present classification results using 
the MNIST data set \cite{MNIST}. Section~\ref{section_discussion} discusses interpretation and implications 
of our findings and also future directions. Appendices provide further technical details.

\section{Data generation and regression}
\label{details_of_cals}

\subsection{Regression setup}

The main idea behind using tensor networks for machine learning is to map features into an exponentially 
high dimensional space and then perform linear regression in that space, but with compressed weights. The weights 
are represented by a tensor with a number of indices in correspondence to the number of inputs. 
This idea was introduced first in Ref.~\cite{stoudenmire2016supervised, Exp_machines}. 
While the amount of memory required to store an arbitrary tensor scales exponentially with the number of its 
indices, compressed tensor representations, called tensor networks, provide a trade-off between 
expressivity versus memory and computational costs required to compute or optimize the tensor. 
Usually, tensor compression improves space and time complexity from exponential to linear in the 
number of features. 

Different tensor networks may be more suitable for particular types of data than others. For example, 
MPS are routinely used to model one-dimensional (1D) systems with short-range correlations, while PEPS \cite{PEPS_classification} 
tensor networks are specifically designed for two-dimensional (2D) systems, and other architectures like tree tensor networks 
\cite{TTN_Stoudenmire, TTN_generative, TTN_quantum} or MERA \cite{MERA_Reyes_Stoudenmire} are useful 
for 1D systems with long-range correlations. 
In this paper we will focus on MPS as the simplest example of a tensor network. 

To set up MPS based regression, we first map each scalar feature $x_i$ into a vector using a feature map $\phi(x_i)$. 
Some common choices of a feature map are trigonometric $\phi(x_i) = (\,\sin(x), \cos(x)\,)$ or polynomial 
\mbox{$\phi(x_i) = (1, x, x^2, \ldots)$}. In this paper we will use a polynomial feature map with a 
dimension $f=3$, namely \mbox{$\phi(x_i) = (1, x, x^2)$}. We then map 
the vector $\x$ of $N$ features into an $f^N$ dimensional space by (formally) taking the tensor product: 
\begin{align}
\x \to \Phi^{s_1 ... s_N}(\x) & = \phi^{s_1}(x_1) \phi^{s_2}(x_2) \cdots \phi^{s_N}(x_N)
\end{align}
where each index $s_j$ takes the values $1,2,\ldots,f$.
Next we define a weight tensor $W$ which we parameterize as a matrix product state (MPS):
\begin{equation}
\label{MPS}
W^{s_1 ... s_N} = \sum_{i_1,i_2, \ldots, i_{N-1}} A^{s_1}_{i_1} A^{s_2}_{i_1 i_2} ... A^{s_{N-1}}_{i_{N-2} i_{N-1}} A^{s_N}_{i_{N-1}},
\end{equation}
where the dimension of the ``virtual",  internal, or bond indices $i_j$ is called the bond dimension of bond $j$.
The maximum over all bond dimensions is referred to as the bond dimension $\chi$ of the MPS as a whole.

The model $f(\x)$ we will use for regression or classification is given by 
contracting the weight tensor $W$ with the feature tensor $\Phi(\x)$ to return a scalar:
\begin{equation}
f(\x) = W \cdot \Phi(\x) = \sum_{s_1, s_2, \ldots, s_N} W^{s_1 ... s_N} \Phi^{s_1 ... s_N}(\x) \label{eqn:model}
\end{equation}
To compute $f(\x)$ efficiently, the local feature maps are contracted one by one with the MPS tensors
$A^{s_n}_{i_{n-1} i_n}$ and similar efficient contractions can be performed to obtain the gradient of each MPS tensor.

A common way to perform a learning task with a model architecture such as (\ref{eqn:model}) is to define a suitable objective or ``loss'' function 
and optimize the weights such that the loss is minimized.
In our calculations we will minimize mean squared error
\begin{equation}
\mathcal{L} = \frac{1}{2T}\sum_{i=1}^{T} \big(f(\x^i) - y^i\big)^2 + \frac{\lambda}{2} |W|^2, \label{eqn:loss}
\end{equation}
where the summation is over $T$ training samples, $y^i$ is a label corresponding to 
the $i$-th feature vector $\x^i$ and the last term is a standard $L_2$ regularization term.

\subsection{Data generation}

To generate complex data whose labels can be exactly predicted by an MPS, we formulate a ``target'' MPS $W_T$
from which we generate labels for our data samples. This target MPS is distinct from the MPS that we
use to parameterize our model Eq. (\ref{eqn:model}).

The steps to generate our training data are:
\begin{enumerate}
\item Sample $N_T$ different feature vectors $\x^i$ from a multivariate Gaussian distribution
\item Apply the map $\Phi(\x^i)$ 
\item Compute $y^i = W_T \cdot \Phi(\x^i)$
\end{enumerate}
The output of this process is a training set $\{ y^i, \x^i \}_{i=1}^{N_T}$ of size $N_T$.

Our choice of local feature map will be \mbox{$\phi^{s_j}(x_j) = (1, x_j, x^2_j)$}. This makes 
the result of contracting a given feature tensor 
with an MPS also a polynomial $\sum \alpha_{p_1 \ldots p_N} (x_1)^{p_1} (x_2)^{p_2} \ldots (x_N)^{p_N}$
for some coefficients $\alpha_{p_1 \ldots p_N}$. 
By tuning the parameters of $W_T$ we can always get the desired coefficients $\alpha_{p_1 p_2 \ldots p_N}$. 
In our experiments we set up an MPS which generates a polynomial whose degree is controlled by the MPS 
bond dimension. We can also control the complexity of the learning task by reducing or increasing the 
effect of higher order terms by varying a single control parameter $\varepsilon$, which is discussed 
below.

We want to construct an MPS returning an n-th order polynomial with reduced effect of higher order terms.
This can be ensured by a particular choice of matrices constituting MPS tensors. Let us consider the 
following structure for the MPS parameterization of $W_T$:
\begin{equation}
\begin{pmatrix}
... \\
... \\ 
\vdots
\end{pmatrix}
\begin{pmatrix}
I \\ M \ x_2 \\ M^2\ x_2^2 \\ \vdots \\ M^{f-1} x_2^{f-1}
\end{pmatrix}
\dots 
\begin{pmatrix}
I \\ M\ x_{n-1} \\ M^2\ x_{n-1}^2 \\ \vdots \\ M^{f-1}\ x_{n-1}^{f-1}
\end{pmatrix}
\begin{pmatrix}
\vdots & \vdots & \dots
\end{pmatrix}, 
\label{parameterization}
\end{equation}
where $f$ is the feature map dimension, the first matrix dimension is $f \times \chi$ ($\chi$ is the bond dimension), and the
last matrix dimension is $\chi \times f$, $I$ is an identity matrix. Here $M$ is a matrix which we will determine below
and  $M^k$ means $M$ to the power of $k$. In the expression (\ref{parameterization}) above, 
we are viewing each MPS tensor $A^{s_j}_{i_{j-1} i_j}$ as a vector of matrices, which is an
equivalent way of thinking of a tensor with three indices: the vector index is the feature index $s_j$ and the matrix indices are
the left and right virtual or bond indices of the MPS.

To obtain an $n$-th order polynomial as our target function, we may choose nilpotent matrices with index $n+1$, meaning $M^{n+1} = 0$. 
For example, any $L$-dimensional triangular matrix with zeros along the main diagonal is nilpotent, 
with index $n \leq L$. This implies that with this choice of $M$-matrices, to get an $n$-th order 
polynomial, the bond dimension must be at least $n+1$. 
For example, to generate the second order polynomial, one may choose the following order $3$ nilpotent $M$ matrix:
\begin{equation}
M = 
\begin{pmatrix}
0 & \varepsilon & 0 \\ 
0 & 0 & \varepsilon \\ 
0 & 0 & 0
\end{pmatrix}, 
M^2 = 
\begin{pmatrix}
0 & 0 & \varepsilon^2 \\ 
0 & 0 & 0 \\ 
0 & 0 & 0
\end{pmatrix}, 
M^3 = 
\begin{pmatrix}
0 & 0 & 0 \\ 
0 & 0 & 0 \\ 
0 & 0 & 0
\end{pmatrix},
\end{equation}
where the value of $\varepsilon < 1$ controls the effect of higher order terms, which means that each 
$n$-th order term has a prefactor $\varepsilon^n$. In general, we can consider more general nilpotent 
matrices $U M U^{\dagger}$, where $U$ is a random unitary matrix.

We noticed that the data generated this way is challenging to fit: the distribution 
of labels $y^i$ is very broad and gets even broader as one includes higher order terms or increases 
their effect. Therefore, to be able to fit the data efficiently, we normalize labels by subtracting their 
mean and dividing by the standard deviation $y^i \to (y^i - \text{mean}(y) ) / \text{std}(y)$. 
Note that despite this post-processing step, one can show that the training data can still be fit perfectly by a model whose
weight MPS $W$ has the same bond dimension as the MPS defining $W_T$.

\subsection{Optimization}

There are multiple strategies to optimize an MPS for a supervised learning task. One method is the density matrix renormalization group (DMRG) algorithm which is the gold standard in physics applications and involves optimizing one or two MPS tensors at a time, sweeping back and 
forth over the MPS \cite{schollwock2011density}. Another option is gradient descent in the style of back-propagation, 
updating every MPS tensor in parallel at each optimization step \cite{Hauru_Riemannian}. 
And there are other alternatives such as Riemannian optimization \cite{Exp_machines}.

We found the DMRG approach to be convenient and efficient, optimizing one MPS tensor at a time 
using conjugate gradient descent based on the \href{https://itensor.org}{ITensor} software \cite{ITensor} and the
\href{https://github.com/Jutho/OptimKit.jl}{OptimKit.jl} library (see details in Appendix~\ref{optimization_appendix_section}).

One issue we had with DMRG, however, was a strong dependence of final results on initial conditions. 
To avoid this issue we developed a second approach based on exact inversion and compression. 
Although the cost of solving for and storing the entire weight tensor $W$ becomes prohibitively large with an increasing number of features, 
and corresponding number of  tensor indices, one can still obtain the weight tensor exactly 
for relatively small problems. To find the exact solution, we set the derivative of the loss, Eq.~(\ref{eqn:loss}), to zero 
which results in linear algebra problem one can solve numerically exactly for $W$.
After finding the exact weight tensor this way, we compress it into an MPS of some desired bond dimension through 
a sequence of singular value decompositions.
When using this inversion and compression method we limited our problem size to only have six features. 
For more details about this method, see Appendix~\ref{exact_inv_mps}. 

While the results obtained via inversion and compression are consistent, meaning they do not depend on any initialization, 
they may be suboptimal as the tensor compression step does not 
guarantee that the resulting MPS is necessarily optimal in the sense of perfectly minimizing the loss. 
Therefore, below we will also present the results of DMRG optimization initialized by the inversion and compression method.

Regardless of the optimization method chosen, the results still depend on the particular training data set used.
Therefore we generate multiple training data sets and a single 
test set from the same probability distribution, run independent calculations for each training set, and average the results 
to obtain the dependence of loss function on the MPS bond dimension with corresponding error bars. 

The source code used for all the experiments in this article can be accessed at \cite{git_code}.

\section{Results with artificial data}
\label{results_with_artificial_data}

In our calculations we use data $\x$ of dimension $N = 6$ and feature map dimension $f = 3$, i.e. each scalar feature 
is mapped into a vector $\phi(x) = (1,x,x^2)$. The target MPS $W_T$ has the maximum possible bond dimension which for $N=6$ and
$f=3$ gives a bond dimension of $\chi=27$ at the center bond. 
Below we present the results of inversion and compression and support them with optimization results, which 
are much more computationally expensive but which can outperform the inversion ones. 

\begin{figure}
\includegraphics[width=0.99\linewidth]{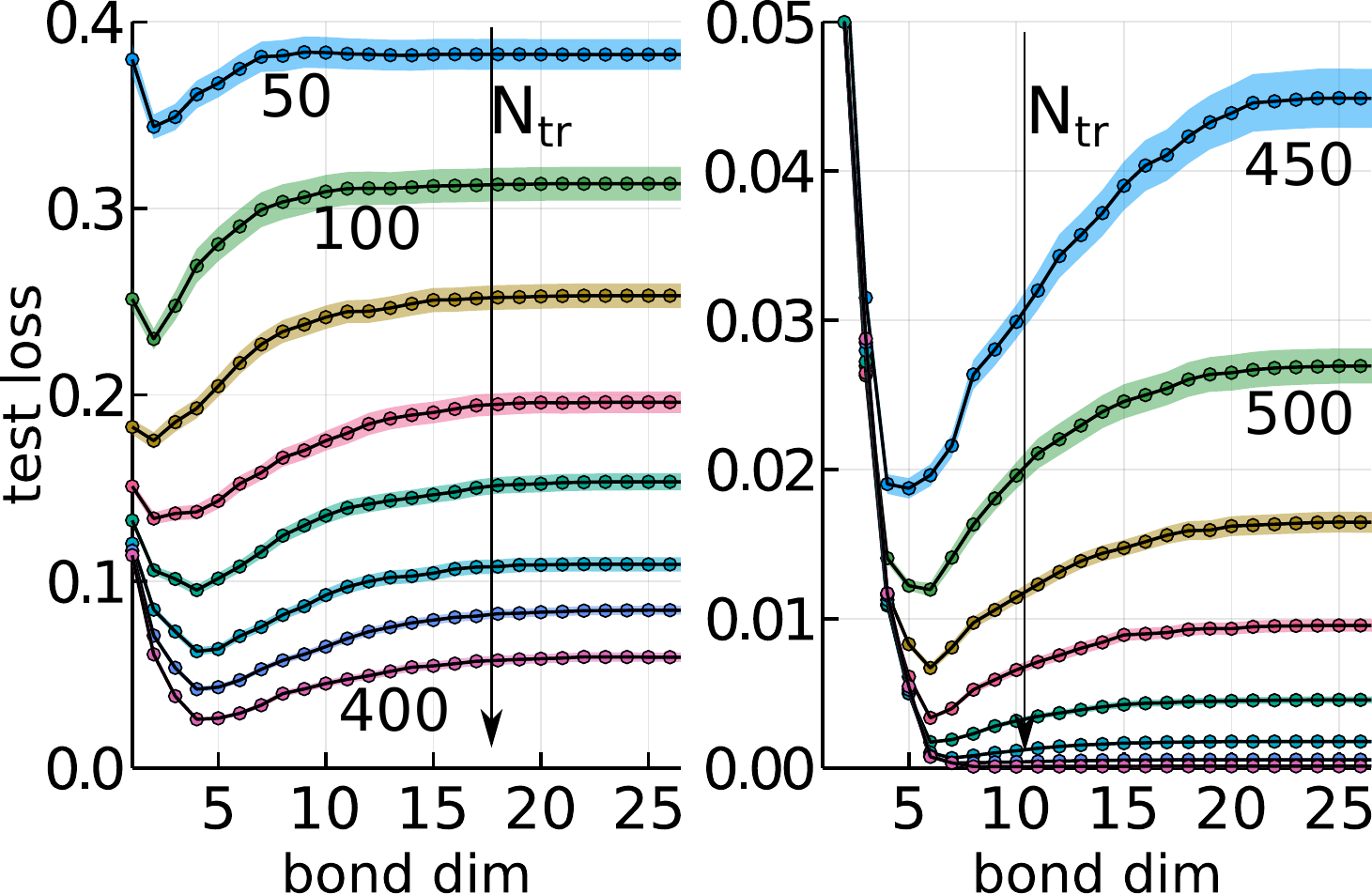}
\caption{Test loss resulting from MPS optimized with the inversion and compression (bond dimension reduction) method. 
        The test set used has $N_{\text{test}} = 1024$ samples.
        The curves show the mean value of test loss and shaded regions show one standard deviation $\pm \sigma$
        when averaging results over $100$ training data sets. 
        The $L_2$ regularization used was $10^{-6}$ and MPS complexity parameter used was $\varepsilon = 0.3$. 
        Different curves correspond to changing the number of $N_{\text{tr}}$ samples in each training set used
         from $50$ to $800$ in steps of $50$ (top to bottom). }
\label{exact_inv_test_loss_different_Ntr}
\end{figure}

\begin{figure}
\includegraphics[width=0.5\linewidth]{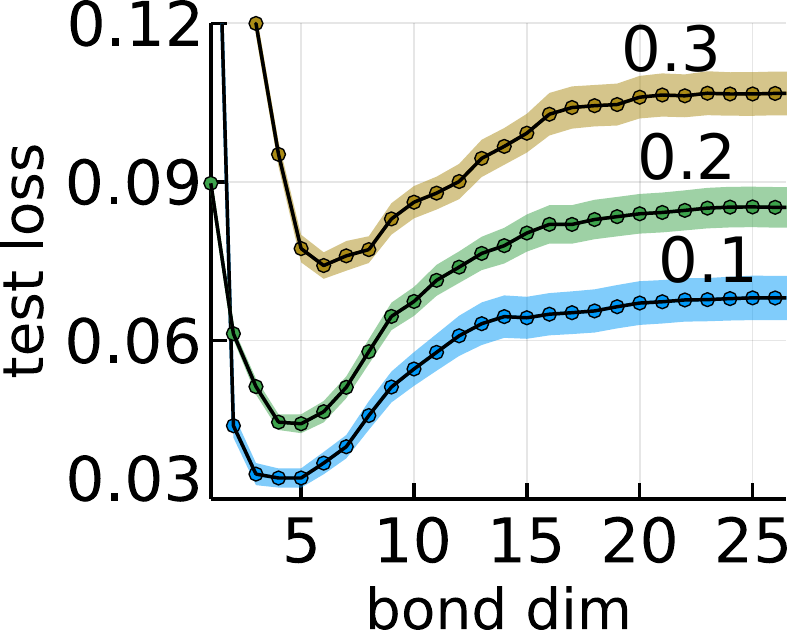}
\caption{Test loss as a function of MPS bond dimension for increasing values of the data complexity parameter $\varepsilon$
         from $0.1$ to $0.3$.
        The inversion and compression method was used to train the model with an $L_2$ regularization of $10^{-6}$. 
        Curves show the mean value and shaded regions one standard deviation when averaged over $100$ data sets. 
        Each training set consisted of $N_{\text{tr}} = 300$ samples and the test set had $N_{\text{test}} = 1024$  samples.
        }
\label{exact_inv_test_loss_different_epsilon}
\end{figure}

In Figures~\ref{exact_inv_test_loss_different_Ntr} and \ref{exact_inv_test_loss_different_epsilon}  
we show the results obtained when using the inversion and compression method to train the model.
Figure~\ref{exact_inv_test_loss_different_Ntr} shows the test loss versus bond dimension as 
the number of training samples increases. 

When the number of training samples is not sufficient to fully fix the weight tensor (which is generally 
the case), we can clearly see a minimum at a bond dimension $\chi^*$, which is much smaller than maximum 
bond dimension. We call $\chi^*$ the optimal bond dimension since it yields the best generalization. 
The optimal bond dimension goes up as we increase the number of training samples. This makes sense as the 
training set becomes more representative of the true data distribution (contains more signal) and the larger 
MPS bond dimension allows a better fit to this data, correspondingly resolving finer details of an underlying distribution.
We also get nearly perfect performance when the number of training samples $N_{\text{tr}} \geq f^N$ 
is sufficient to fully characterise the weight tensor, which happens when the number of samples reaches $f^N$ ($= 3^6 = 729$ in our case).

Figure~\ref{exact_inv_test_loss_different_epsilon} shows the evolution of the test loss 
dependence of an MPS bond dimension as the effect of higher order terms varies by means of changing 
data generating MPS control parameter $\varepsilon$. 
As $\varepsilon$ increases, the effect of higher order terms becomes more pronounced and the data 
essentially becomes more complex. This leads to the optimal bond dimension $\chi^*$ being 
shifted towards higher values required to resolve more delicate details of data. 

While the procedure described above is numerically cheap and efficient, it does not guarantee that 
the results are optimal for all bond dimensions, only for the largest bond dimension. This is because the tensor compression
step of our inversion and compression procedure only guarantees that the MPS we obtain 
is the closest MPS to the original, optimal tensor in $L_2$ norm, which is of course different from minimizing the loss directly. 
To verify the robustness of our conclusions, we now turn to DMRG optimization, which directly minimizes the loss at a fixed bond dimension but is more numerically costly in our case.

To verify that we obtain similar results when using DMRG optimization to train our model,
we focus on a particular training set size $N_{\text{tr}} = 300$ and target model complexity $\varepsilon = 0.1, 0.3$ 
(which is small enough reflecting our assumption that the data can be efficiently modeled by a low order polynomial of 
features thus making an MPS a good enough weight ansatz for such purposes). We sample several 
training sets and and run both inversion and compression as well as DMRG optimization on each of them.
Results for other training data set sizes and complexities $\varepsilon$ are shown in Appendix~\ref{supplem:inv_compress_and_optim_for_more_data_sets} and qualitatively agree with the results presented in this section.

\begin{figure}
\includegraphics[width=0.99\linewidth]{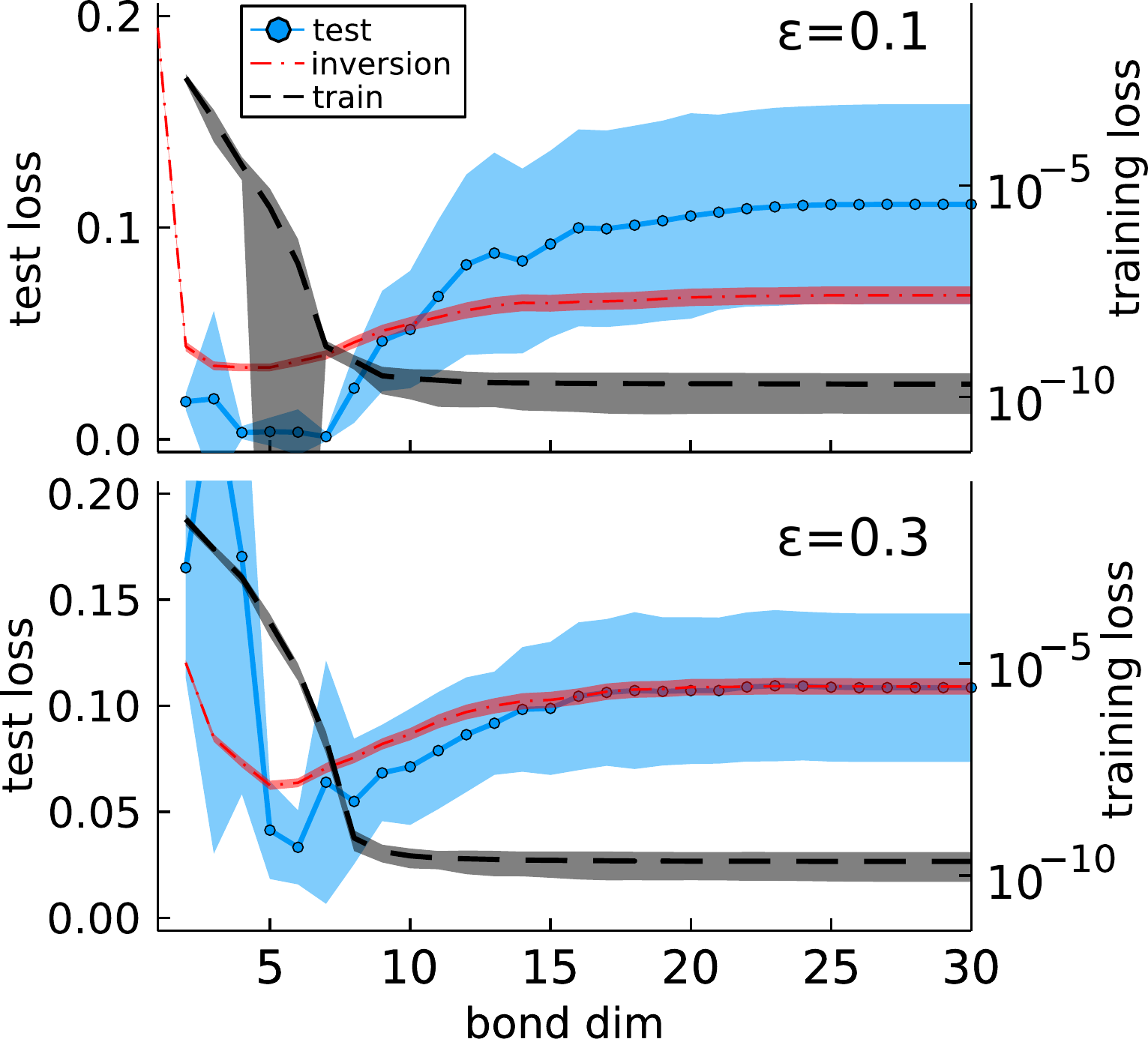}
\caption{The results of DMRG optimization at various MPS bond dimensions averaged over 32 training data sets each of
        size  $N_{\text{tr}} = 300$. The complexity parameter of the target distribution was  $\varepsilon = 0.1, 0.3$ (top and bottom panels respectively).
        The test set size used was $N_{\text{test}} = 1024$.
        The dash--dotted red line and the corresponding shaded region show 
        the mean test loss obtained using the inversion and compression method with one sigma standard 
        deviation. The blue line (with shaded regions corresponding to one sigma standard deviation) shows the 
        mean value of test loss at the model state corresponding to the best 
        validation loss. The black dashed line shows mean training loss (and shaded region showing one standard deviation). 
        }
\label{optim_results}
\end{figure}

In Fig.~\ref{optim_results} we show the resulting loss values obtained with both optimization methods.
We can see that overall the inversion and compression approach returns very good results despite not involving the loss
function in the compression step of the procedure.
Only in the vicinity of optimal bond dimension does additional DMRG optimization consistently improve model performance. 
As in the other results above, we can clearly see 
that the test loss starts degrading (increasing) beyond a certain bond dimension while the training loss continues to improve.
Overall, our DMRG optimization results confirm what we observed with the inversion and compression method: that overfitting
occurs beyond a certain bond dimension $\chi^*$ whose value depends on the size of the training set and complexity of the 
underlying data distribution.

\section{Results with MNIST}
\label{results_with_MNIST}

The results obtained with artificial data show that the best performance on unseen test data happens for a finite
MPS bond dimension, suggesting that limiting the bond dimension may be a good strategy to improve generalization and prevent overfitting.
In this section we check whether this conclusion holds for the task of supervised learning of the MNIST data set of handwritten digits. 
For these experiments we will be using cross-entropy loss function, discussed in 
Appendix~\ref{supplem:classification_appendix_section}.

When using the full training set of MNIST images, we see that training accuracy steadily improves as the bond 
dimension grows, although in our experiments the training accuracy never reached hundred per cent for the full 
MNIST training set (not shown). Inspired by Ref.~\cite{double_descent_Sutskever}, which shows that the 
second descent of training loss occurs when perfect training accuracy is reached,
we reduce the number of training images to $1024$ so that we can attain this limit.

\begin{figure}
\includegraphics[width=0.99\linewidth]{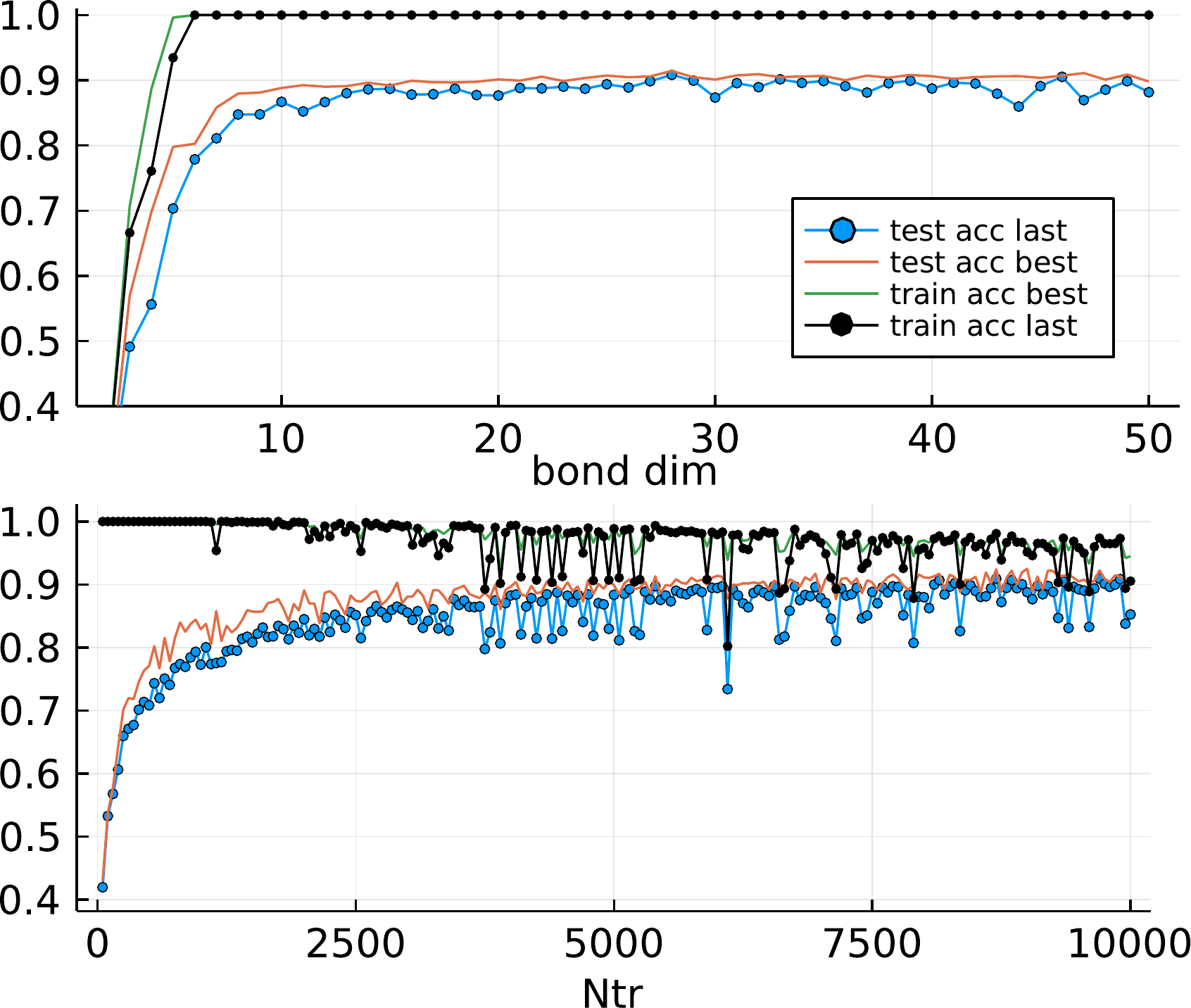}
\caption{The results of optimization obtained with a sweeping algorithm at fixed bond dimension for 
        an MNIST data set. Top panel shows the best test loss and training and test 
        losses at the last training step (100 full DMRG sweeps --- from left to right 
        and right to left --- with 5 conjugate gradient steps per MPS tensor) using $1024$ training images.
        Bottom panel shows losses versus the number of training images using an MPS of fixed bond dimension 
        $\chi = 6$.
        }
\label{loss_vs_Ntr_and_chi_MNIST}
\end{figure}

In the top panel of Fig.~\ref{loss_vs_Ntr_and_chi_MNIST} we plot 
training and test set accuracies versus MPS bond dimension at a fixed training set size $N_{\text{tr}} = 1024$ and test 
set size $N_{\text{test}} = 10,000$. The model reaches perfect training accuracy at bond dimension $\chi = 6$ and we do not 
see any test accuracy drop or recovery in the vicinity of $\chi = 6$. 
Overall, the test set accuracy only continues to improve as the bond dimension is increased and we do not see any indication of double 
descent behavior (even when introducing label noise as proposed in Ref.~\cite{double_descent_Sutskever} --- 
see Fig.~\ref{MNIST_with_noise_long_run} in Appendix~\ref{supplem:MNIST_with_noise_long_run}). Moreover, 
we also do not see any overfitting of the kind we observed in our artificial data experiments. In other words, we essentially see a single descent. 
We note that it could be possible that as we  
increased the bond dimension $\chi$, we simply overshot a small range of bond dimensions where the test set accuracy
would show a double-descent feature. One reason this could be possible is that the number of parameters in our model grows quadratically with $\chi$, so rather quickly.

To further investigate whether overfitting or double descent might be possible, 
we follow a procedure similar to one in Ref.~\cite{double_descent_linear_regr} 
by fixing MPS bond dimension $\chi = 6$ and varying the number of training samples. In the lower panel of 
Fig.~\ref{loss_vs_Ntr_and_chi_MNIST} we plot corresponding accuracies. 
Although the results are a bit noisy, overall they show monotonic improvement of test set accuracy 
(and degrading training accuracy) as the number of training samples increases. 
This is in contrast with the results of Ref.~\cite{double_descent_linear_regr}, which show strongly 
degraded test performance in the vicinity of $N_{\text{tr}}$, which corresponds to transition between 
perfect and imperfect fitting of training data, which would happen around $N_{\text{tr}} \approx 1000$ 
in our case. 

Therefore, surprisingly with the MNIST data as we increase the MPS bond dimension we neither see overfitting, 
which we see with our artificial data, nor do we see double descent
 which has been observed in many other machine learning architectures. 
 One possibility is that overfitting 
or a double descent simply occurs at much larger bond dimensions than we reached. 
But this is unlikely as in the top panel of Fig.~\ref{loss_vs_Ntr_and_chi_MNIST} 
we pushed the bond dimension far beyond one required 
for perfect interpolation, that is, achieving zero training error. 
Another explanation could come from the essentially 
different nature of the artificial and MNIST data sets. The artificial data is in some sense one-dimensional as it is created by an MPS and  
can be exactly fitted by an MPS, while MNIST data is two-dimensional, for which MPS is not ideally suited in principle. 
Indeed, as was shown in Ref.~\onlinecite{PEPS_classification}, two-dimensional PEPS tensor network architectures do show overfitting with MNIST 
and quasi-two-dimensional block-MPS architectures in Ref.~\onlinecite{block_MPS} show overfitting as well.
Therefore, we speculate that the fitting and generalization properties of tensor networks not only differ from other machine 
learning architectures, but also depend on the data itself. 

\begin{figure}
\includegraphics[width=0.99\linewidth]{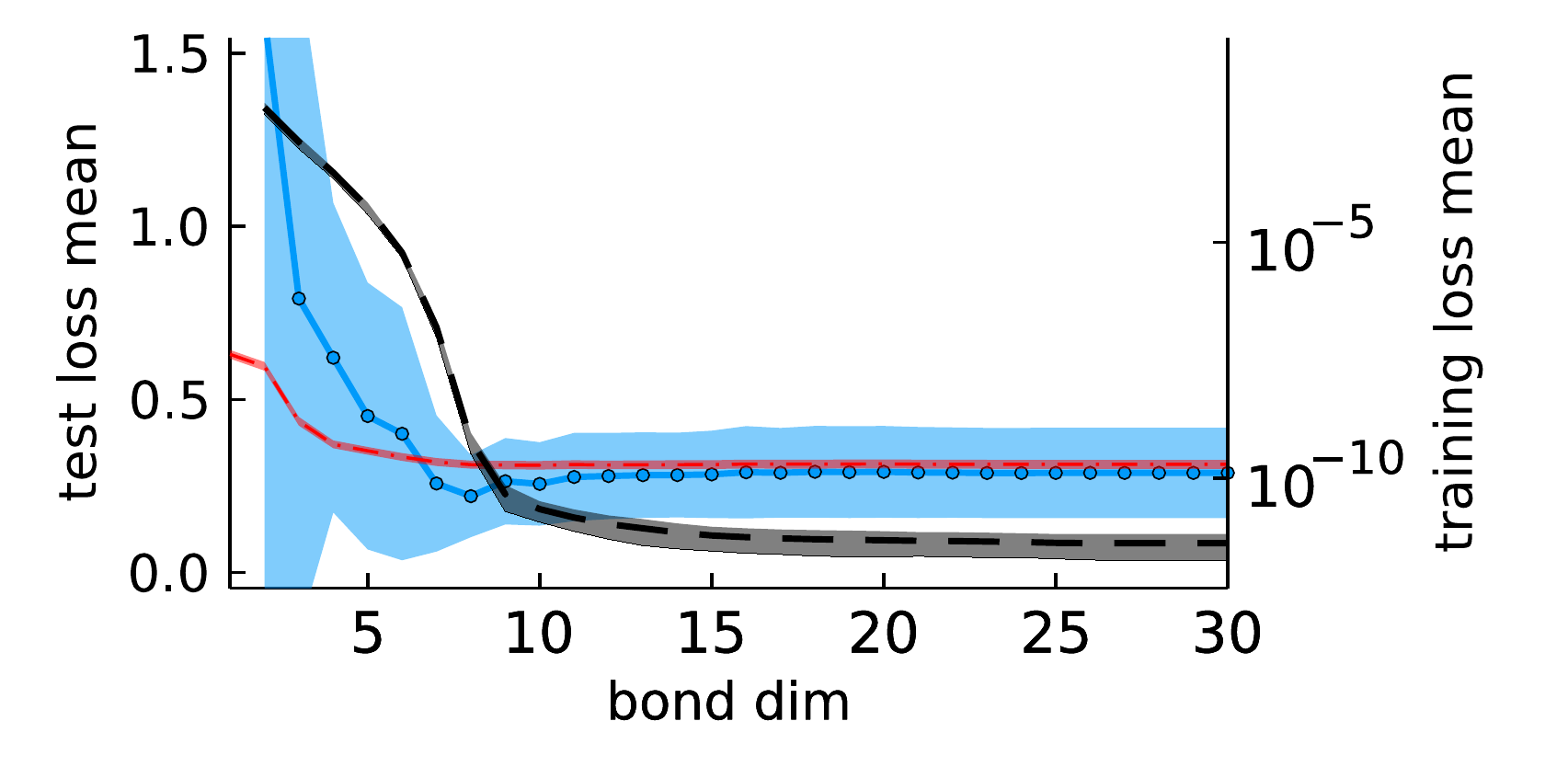}
\caption{The results of DMRG optimization at various MPS bond dimensions averaged over 32 training data sets each of
        size  $N_{\text{tr}} = 300$. The complexity parameter of the target distribution was  $\varepsilon = 1.0$. 
        All other parameters and lines are as in Fig.~\ref{optim_results}.
        }
\label{losses_eps_1.0}
\end{figure}

To illustrate the last point, we construct an artificial dataset with $\varepsilon = 1.0$, therefore making the data too complex 
to be efficiently fit by an MPS of a fixed (not exponentially large) bond dimension. In Fig.~\ref{losses_eps_1.0} we show 
the results of inversion with compression and DMRG optimization. 
We can see that with more complex data overfitting is much weaker and so increasing the bond dimension 
does not lead to pronounced U-shape behaviour of test loss as in Fig.~\ref{optim_results} for less complex data. 
This further suggest that generalization properties of tensor networks can depend on the combination of their architecture 
and the properties of the data.

\section{Discussion}
\label{section_discussion}

We studied the generalization properties of MPS based models using artificial data and as well as more realistic MNIST image data. 
Our results with artificial data show overfitting for large enough MPS bond dimensions $\chi$, 
such that limiting $\chi$ to a finite value typically improves model performance on unseen test data. 
In contrast, we found for MNIST that increasing $\chi$ always improved test performance, leading to 
a ``single descent" and no overfitting. 

We did not find any signatures of the double descent 
behavior frequently observed in other model architectures, such as in Refs.~\onlinecite{double_descent_Sutskever, double_descent_linear_regr}. 
Of course there is no guarantee that the MPS regression approach should exhibit double descent, 
as it is different in many respects from other established machine learning techniques. But the presence or absence of the
double descent phenomenon should be investigated further for MPS using other data sets and larger bond dimensions.
The same line of investigation should also be pursued for more expressive types
of tensor networks such as tree tensor networks \cite{TTN_Stoudenmire,TTN_generative}, block MPS 
\cite{block_MPS}, or PEPS \cite{PEPS_classification} The generalization and overfitting behavior of other features maps used in conjunction with MPS should also be investigated, such as feature maps based on data ``amplitude encoded'' as MPS \cite{Wright_Deterministic,Dilip:2022}.

One lesson highlighted by our observation of overfitting is the very different settings of training an MPS on a finite amount of 
data versus optimizing an MPS for a fully specified problem. 
Historically, the most common application of MPS has been for solving partial differential equations,
most notably the Schr\"odinger equation in quantum physics. 
For such problems, increasing the bond dimension can always yield more accurate results because the 
differential operator is exactly specified. 
But when using MPS to model sampled data, our results show that continuing to increase the bond dimension 
may not always lead to better outcomes due to the statistical variance inherent to a training set of a finite size.

Another issue we encountered in our work is that training tensor network models can be difficult in certain cases.
For example, in Ref.~\cite{Barren_plateaus_TNs}  it was shown that barren plateaus (regions of an exponentially vanishing gradient and thus inefficient training) are very common with MPS when using a global loss function. On the other hand, MPS training on the MNIST is very efficient, 
while on the artificial data (which can be fit exactly by an MPS) training turned out to be much more challenging, 
which suggests that trainability depends not only on the type of model, but on the data itself. 

Finally, our differing results for the artificial data versus MNIST image data highlights the need to study what kind of 
data can be efficiently modeled by different kinds of tensor networks. Work along these lines is  
already under way \cite{Information_theor_TNs,MPS_for_classical_data,block_MPS}
but open questions remain. For example, while exact calculations show that MPS cannot efficiently capture certain distributions 
of two-dimensional image data \cite{block_MPS}, approximating these same distributions with finite bond dimension 
MPS nevertheless yields good unsupervised and supervised model performance \cite{MPS_without_optimization,Wright_Deterministic}.
Understanding how MPS and other tensor networks capture data sets will not only lead to better practices for
machine learning, but could also spur progress toward more ambitious goals such as a classification of data.

\begin{acknowledgements}

EMS would like to acknowledge David Schwab for insightful discussions on the topic of overfitting with tensor networks. The Flatiron Institute is a division of the Simons Foundation. 

\end{acknowledgements}

\clearpage
\bibliographystyle{apsrev4-1}
\bibliography{literature}

\onecolumngrid
\clearpage

\appendix

\section*{Supplementary Material for ``MPS generalization: bond dimension effect''}
\twocolumngrid

\section{Inversion and compression}
\label{supplem:inversion_and_compression}

We want to minimize the loss function 
\begin{equation}
\mathcal{L} = 
    \frac{1}{2T} \sum_{i=1}^T 
    \left( \Phi_{\xi}(\x^i) W_{\xi} - y^i \right)^2 
    + 
    \frac{\lambda}{2} W_{\xi} W_{\xi}
\end{equation}
over MPS tensors $W_{\xi}$. The derivative of the loss function over $W_{\eta}$ is 
\begin{equation}
\label{gradient}
\frac{\partial \mathcal{L}}{\partial W_{\eta}} = 
    \frac 1 T \sum_{i=1}^T 
    \Phi_{\eta}(\x^i) \left( \Phi_{\xi}(\x^i) W_{\xi} - y^i \right) + \lambda W_{\eta}. 
\end{equation}
Requiring the first derivative to be zero and reorganizing the terms, we get the following equation for 
the weight tensor $W$:
\begin{equation}
\label{exact_inv_mps}
 A_{\alpha \eta} W_{\eta} = b_{\alpha},
\end{equation}
where 
\begin{equation}
A_{\alpha \eta } = 
\left[
    \lambda \delta^{\alpha \eta } + \frac 1 T \sum_{i=1}^T \Phi_{\alpha}(\x^i) \Phi_{\eta}(\x^i)
\right]
\end{equation}
and 
\begin{equation}
b_{\alpha} = \frac 1 T \sum_{i=1}^T y^i \Phi_{\alpha}(\x^i).
\end{equation}

Therefore, we need to invert an $f^N \times f^N$ matrix $A_{ \alpha\eta}$, where $N$ is the number of features. 
Thus, if the number of independent training samples $N_{\text{tr}}$ is larger or equal to $f^N$, 
the $A$ matrix has full rank and a unique solution exists. If $N_{\text{tr}} < f^N$, the kernel space is 
not empty and infinitely many solutions exist and exact inversion is ill--defined. To overcome this 
issue, we introduce small but finite regularization parameter $\lambda$, which guarantees that the 
matrix $A$ is full rank. For inversion we then use a ``backslash" method of the standard Julia linear algebra 
package (based on LU factorization). Once the exact tensor is obtained, we then 
compress it by constructing MPS of a given bond dimension using a sequence of singular value decompositions as
implemented the ITensor library \cite{ITensor}.

\section{Optimization Procedure}
\label{optimization_appendix_section}

In our calculations we implemented the simplest optimization approach based on optimization of each 
MPS tensor one by one, one tensor at a time. This approach is similar to one outlined in 
Ref.~\cite{stoudenmire2016supervised}, but is simpler as one optimizes a single tensor rather 
than a double contracted tensor with subsequent SVD and dropping small singular values. 
We found that at small bond dimensions, DMRG-like approach retaining only dominant singular 
values can be sub-optimal as neglecting smaller singular values may have detrimental 
effect on loss function if bond dimension is relatively small. 

For optimization we used the ``blisfully ignorant" \href{https://github.com/Jutho/OptimKit.jl}{OptimKit.jl} 
library, which offers powerful optimization routines, including conjugate gradient descent, and requires 
one only to provide the objective function, its gradient, and the initial conditions, while not requiring any 
particular choice of data types. This setup is very convenient as it allows one to optimize MPS tensors 
directly without intermediate  conversion to e.g.\ standard 1D arrays.

\section{Classification}
\label{supplem:classification_appendix_section}

For classification our approach is similar to the one in Ref.~\cite{stoudenmire2016supervised} with 
the only difference that rather than using the mean square error, here we used cross entropy loss function, 
which is basically 
\begin{equation}
\label{cross_entr}
- \frac 1 T \sum_{i=1}^T \ln p_i^{\ell} \delta_{\ell, \tilde{\ell}_i},
\end{equation}
where $p_i^{\ell}$ is the resulting ``probability" of the i-th image belonging to the class $\ell$, 
$\tilde{\ell}_i$ is a known class which an $i$-th image belongs to, and $T$ is the total number 
of training images. 
While $p_i^{\ell}$ does not necessarily reflect actual classification probability due to calibration issues --- 
see, e.g., Ref.~\cite{Neural_Nets_Calibration}, cross entropy is still a reasonable and widely used 
choice of a loss function for classification. 

To obtain $p_i^{\ell}$, we can use an MPS with just an extra open index $\ell$, which essentially 
means that we add one more index to one of $A$ tensors of an MPS in Eq.~\ref{MPS}. Contraction of an MPS with an 
image tensor results in a vector rather than a scalar as in Eq.~\ref{eqn:model}. 
Squaring and normalising the resulting vector elements returns numbers, which can be interpreted as 
probabilities of an image belonging to each class $\ell$ and used in Eq.~\ref{cross_entr}.

\section{Optimization and inversion for other artificial data sets}
\label{supplem:inv_compress_and_optim_for_more_data_sets}

In this section we show the results of optimization and inversion similar to ones in 
Figs.~\ref{exact_inv_test_loss_different_epsilon} and \ref{optim_results}, but for a wider range of 
the parameter $\varepsilon$, which controls the effect of higher order terms. 

\begin{figure}
\includegraphics[width=0.99\linewidth]{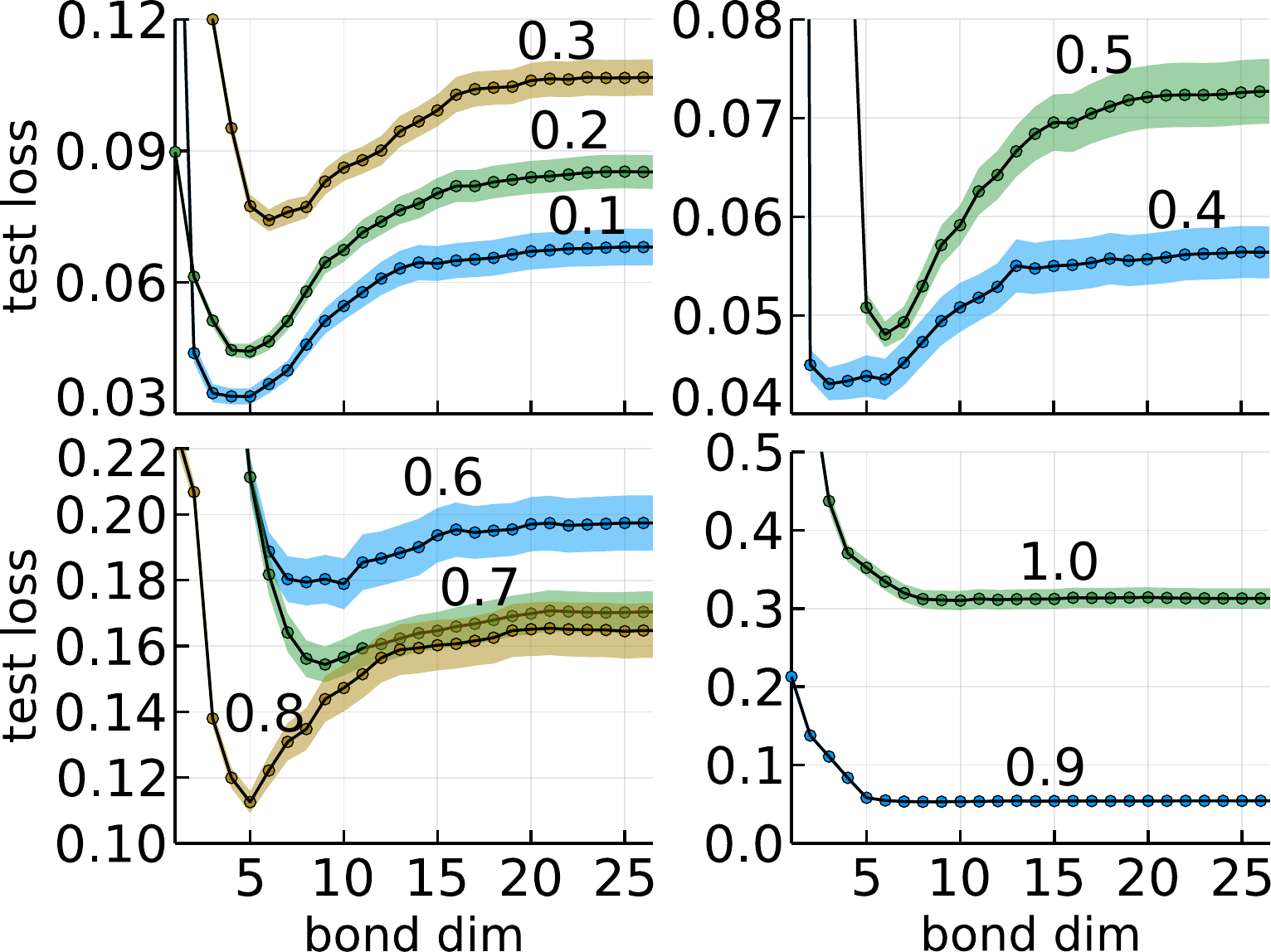}
\caption{Inversion with subsequent tensor compression (bond dimension reduction). 
        Test loss. Mean value and $\pm \sigma$. Averaged over $100$ data sets. 
        $L_2$ regularization $10^{-6}$. 
        Changing the MPS parameter $\varepsilon$ from $0.1$ to $1.0$ 
        using $N_{\text{tr}} = 300$ training samples.}
\label{exact_inv_test_loss_different_epsilon_full}
\end{figure}

Figure~\ref{exact_inv_test_loss_different_epsilon_full} extends the results of inversion with subsequent 
tensor compression shown in Fig.~\ref{exact_inv_test_loss_different_epsilon}, to larger values of $\varepsilon$. 
We can see that for many values of the parameter $\varepsilon = 0.1, 0.2, 0.3, 0.5, 0.6, 0.7$, 
which controls the effect of higher order terms, the results follow the general trend discussed in the 
main text: the optimal bond dimension $\chi^*$ shifts towards higher values as the effect of higher order 
terms increases. However, surprisingly we find that in our case the results obtained with 
$\varepsilon = 0.4, 0.8$ do not follow this simple rule. Moreover, for $\varepsilon = 0.9,1.0$ --- 
when the data is not 1D in a sense that it cannot be efficiently modeled by an MPS (with restricted bond dimension) 
due to highly non-local and high-order correlations --- the results of inversion and compression suggest that 
that the test loss monotonically decreases with bond dimension and saturates at higher bond dimension without any signature of 
overfitting. 
This is similar to our results with MNIST data set shown in top panel of Fig.~\ref{results_with_MNIST}: 
the test accuracy improves as bond dimension grows and then saturates, but does not degrade as bond 
dimension grows even further.

\begin{figure}
\includegraphics[width=0.99\linewidth]{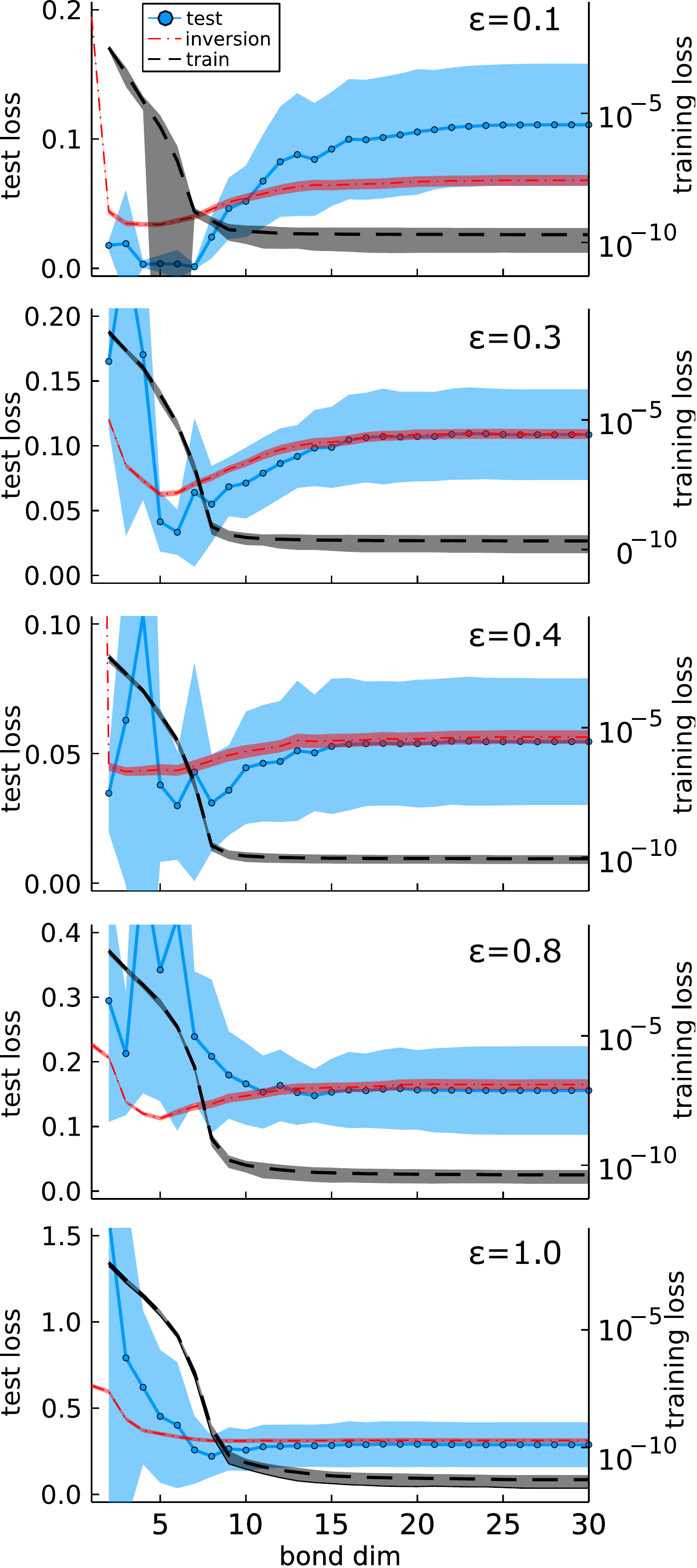}
\caption{This figure extends the results presented in Fig.~\ref{optim_results}: 
        it shows test (dotted blue line) and training (black dashed line) losses obtained with the DMRG optimization 
        approach and also test loss (dot--dashed red line) obtained with the inversion and compression approach for a wider 
        range of data complexities characterised by $\varepsilon$.}
\label{optimization_different_eps}
\end{figure}

Figure~\ref{optimization_different_eps} extends the results of inversion with subsequent 
tensor compression shown in Fig.~\ref{optim_results}, to other values of $\varepsilon$. 
In general, we find that at large bond dimension the results of inversion and compression 
match optimization results well, while at smaller bond dimensions the match is more qualitative: 
although in most of the cases both methods result in U-shape behaviour of test loss, the actual values of 
test loss are notably different.
At small values of $\varepsilon$, DMRG optimization generally leads to better 
test results at smaller bond dimensions, while it is not always true for larger $\varepsilon$. 

Overall, these results support the claim that reducing bond dimension can lead to better 
test performance, although they also highlight the challenges and open questions associated with the effect of data 
properties on MPS optimization and its performance.

\section{Longer optimization for MNIST classification with training label noise}
\label{supplem:MNIST_with_noise_long_run}

While the results presented in Fig.~\ref{loss_vs_Ntr_and_chi_MNIST} do not show any evidence of 
double descent behavior, it is possible that running calculations much longer with label noise 
(randomly corrupting some portion of training labels) 
may reveal the second descent. Indeed, the results presented in e.g., Ref.~\cite{double_descent_Sutskever} 
were obtained using a few thousands of optimization steps and also with some finite level of 
label noise (without noise in some cases the double descent is hardly notable and we may just 
overshoot it by increasing MPS bond dimension).

\begin{figure}
\includegraphics[width=0.99\linewidth]{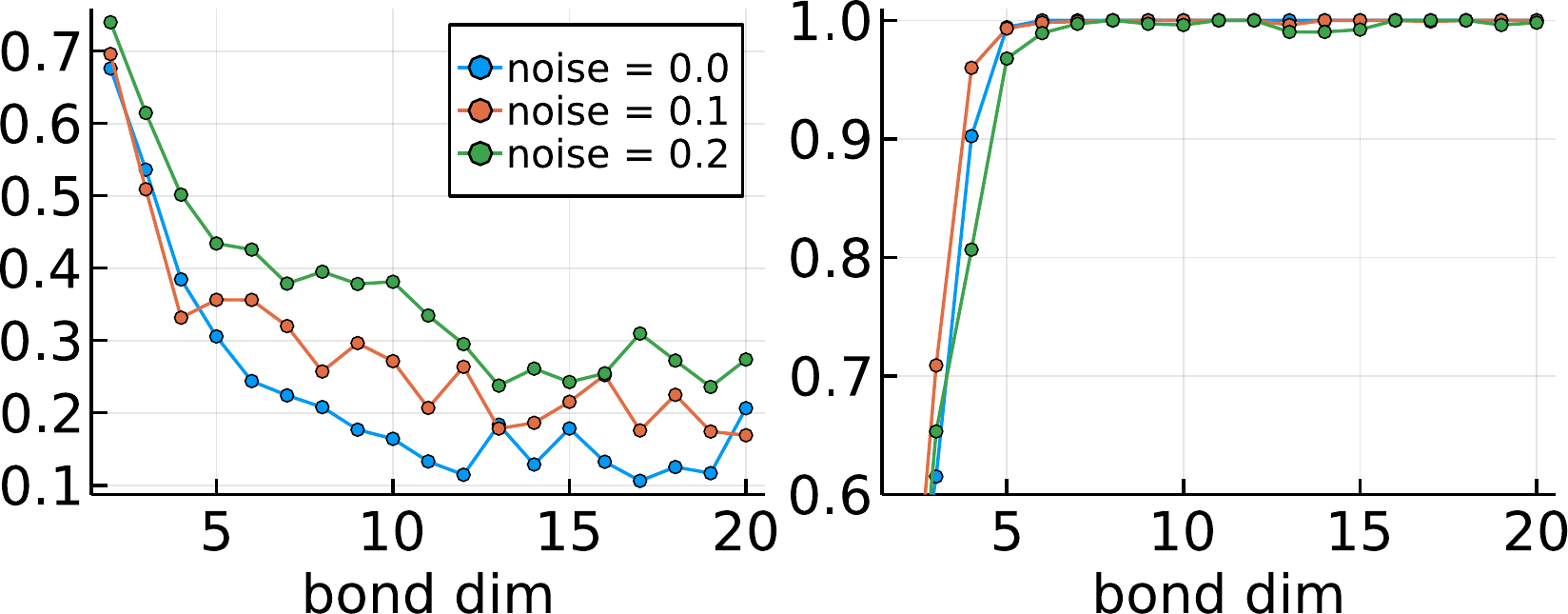}
\caption{Model performance on MNIST. $N_{\text{tr}} = 1024$ training images and 
            $0,10,20\%$ label noise level for up $2000$ epochs. Left panel shows 
            test error vs bond dimension at different level of label noise (the fraction 
            of training images with randomly corrupted labels). Right panel shows 
            training accuracy vs bond dimension.}
\label{MNIST_with_noise_long_run}
\end{figure}

In Fig.~\ref{MNIST_with_noise_long_run} we show the results of calculations with MNIST using 
$N_{\text{tr}} = 1024$ training images and $0,10,20\%$ label noise level for up $2000$ epochs 
(5 days of calculations, which resulted in $800-2000$ epochs depending on noise and bond dimension), 
for $\chi \in [2,20]$. We can see that even with noise, increasing bond dimension generally leads 
to better test performance and adding noise does not lead to any double descent behavior. 

\clearpage

\end{document}